\documentclass[runningheads]{llncs}
\usepackage{graphicx}

\usepackage{tikz}
\usepackage{comment}
\usepackage{amsmath,amssymb} 
\usepackage{color}

\usepackage[accsupp]{axessibility}  

\usepackage[skip=2pt]{caption}
\usepackage{algorithm,algpseudocode,mathrsfs,subcaption}
\usepackage{subcaption}
\usepackage{bm}
\usepackage{graphicx}
\usepackage{enumitem}
\usepackage{multicol}
\usepackage{multirow}
\usepackage{xspace}
\usepackage{amsfonts}

\usepackage{wrapfig,lipsum,booktabs}
\usepackage{array}
\usepackage{changepage}

\newcommand\ChangeRT[1]{\noalign{\hrule height #1}}
\def\eg{\emph{e.g.,}~} 
\def\ie{\emph{i.e.,}~}

\def\x{\mathbf{x}}

\def\v{\mathbf{v}}
\def\g{\mathbf{g}}
\def\a{\mathbf{a}}
\def\tb{\bm{\theta}}

\begin{document}
\pagestyle{headings}
\mainmatter
\def\ECCVSubNumber{7022}  

\title{Revisiting Outer Optimization in Adversarial Training} 

\titlerunning{Revisiting Outer Optimization in Adversarial Training}
%
\author{Ali Dabouei \and
Fariborz Taherkhani \and
Sobhan Soleymani \and
Nasser M. Nasrabadi}
\authorrunning{A. Dabouei et al.}
%
\institute{West Virginia University\\
\email{\{ad0046, ft0009, ssoleyma\}@mix.wvu.edu, nasser.nasrabadi@mail.wvu.edu}}
\maketitle

\begin{abstract}
Despite the fundamental distinction between adversarial and natural training (AT and NT), AT methods generally adopt momentum SGD (MSGD) for the outer optimization. This paper aims to analyze this choice by investigating the overlooked role of outer optimization in AT. Our exploratory evaluations reveal that AT induces higher gradient norm and variance compared to NT. This phenomenon hinders the outer optimization in AT since the convergence rate of MSGD is highly dependent on the variance of the gradients. To this end, we propose an optimization method called ENGM which regularizes the contribution of each input example to the average mini-batch gradients. We prove that the convergence rate of ENGM is independent of the variance of the gradients, and thus, it is suitable for AT. We introduce a trick to reduce the computational cost of ENGM using empirical observations on the correlation between the norm of gradients w.r.t. the network parameters and input examples. Our extensive evaluations and ablation studies on CIFAR-10, CIFAR-100, and TinyImageNet demonstrate that ENGM and its variants consistently improve the performance of a wide range of AT methods. Furthermore, ENGM alleviates major shortcomings of AT including robust overfitting and high sensitivity to hyperparameter settings. 

\end{abstract}

\section{Introduction}
\label{sec:intro}

\begin{wrapfigure}[10]{r}{0.45\textwidth}
\vspace{-20pt}
    \centering
    \includegraphics[width=0.45\textwidth]{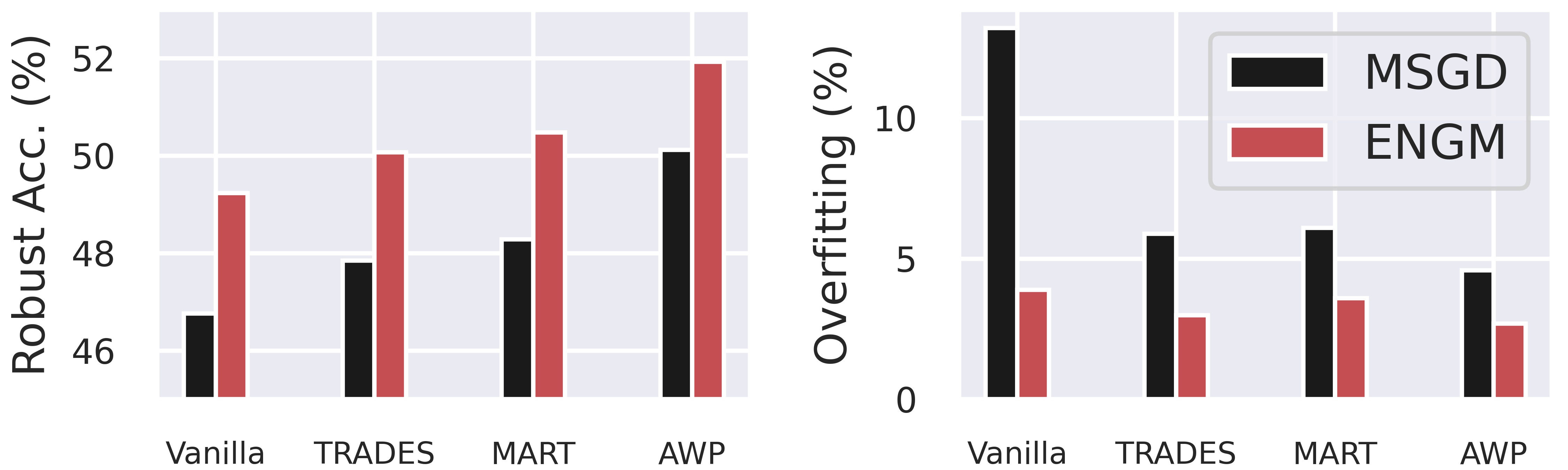}
    \caption{Replacing MSGD with ENGM for outer optimization in AT results in consistent improvement of robust accuracy and generalization. }
    \label{fig:my_label}
\end{wrapfigure}

Susceptibility of deep neural networks (DNNs) to manipulated inputs has raised critical concerns regarding their deployment in security-sensitive applications \cite{chen2015deepdriving,kurakin2016adversarial,ma2021understanding}. The worst-case manipulation can be characterized by {\it adversarial examples}: carefully crafted input examples that can easily alter the model prediction while remaining benign to the human perception \cite{szegedy2013intriguing,goodfellow2014explaining}. A principal approach to formalize the imperceptibility is to bound the perturbation using $\ell_p$-norm. Hence, the problem of finding a model robust to adversarial manipulation reduces to finding the one that generalizes well merely on the bounded neighborhood of the input example. Although this task seems effortless for humans, achieving such invariance is notoriously difficult for DNNs. The reason for this behavior has not been fully understood yet, but several factors have shown to be influential, including the high cardinality of the data space and non-zero test error of the classifier on noisy inputs \cite{ford2019adversarial,cohen2019certified}.



One of the most effective methods (defenses) to alleviate adversarial susceptibility is adversarial training (AT) which improves the robustness by training the model on the worst-case loss \cite{goodfellow2014explaining,madry2017towards}. Given the deep model $F_{\tb}$ parameterized by $\tb$ and the surrogate loss function for the empirical adversarial risk $L$, the training objective of AT is defined as:
\begin{subequations}
\begin{equation}\label{eq:oo} 
     \min_{\tb} \mathbb{E}_{(\x, y)\sim \mathcal{D}}\bigg[ L^*\big(\x, y;\tb\big)\bigg],
\end{equation}
\begin{equation}\label{eq:io} 
     L^*\big(\x, y;\tb\big) =   \max_{||\x-\x'||_p \leq \epsilon} ~~L\big(F_{\tb}(\x'), y\big),
\end{equation}
\end{subequations}
where the input example $\x$ and the corresponding label $y$ are a sample from the data distribution $\mathcal{D}$, $\x'$ is the adversarial equivalent of $\x$, and $\epsilon$ is the maximum $\ell_p$-norm magnitude of the perturbation. 
Concretely, adversarial training consists of two simultaneous optimizations, referred to as the inner and outer optimizations. The inner optimization (Equation \ref{eq:io}) finds the worst-case adversarial example, and the outer optimization (Equation \ref{eq:oo}) minimizes the empirical adversarial risk over the network parameters, $\tb$. 


Numerous efforts have been devoted to analyzing different aspects of AT, such as the inner optimization \cite{madry2017towards,zhang2020attacks,sitawarin2020improving,dabouei2020smoothfool,dabouei2019fast}, adversarial objective \cite{zhang2019theoretically,wang2019improving,pang2021adversarial,dabouei2020exploiting}, computational cost \cite{shafahi2019adversarial,zheng2020efficient,wong2020fast}, and evaluation methods \cite{carlini2017towards,moosavi2016deepfool,athalye2018obfuscated,dong2018boosting,croce2020reliable}. Recent studies on the topic have revealed two major shortcomings of AT which contradicts common observations on NT. First, AT severely induces overfitting \cite{rice2020overfitting,chen2020robust}, referred to as {\it robust overfitting}, whereas in NT overfitting is known to be less prominent especially in over-parameterized models \cite{zhang2017understanding,neyshabur2017exploring,belkin2019reconciling}. Second, AT is highly sensitive to hyperparameter setting, \eg a slight change in the weight decay can deteriorate the robust performance \cite{gowal2020uncovering,pang2020bag}.




The majority of the previous works on AT have analyzed the inner optimization and its properties. However, the potential impact of outer optimization on the performance and shortcomings of AT has been critically overlooked. Furthermore, the success of the two recent state-of-the-art (SOTA) approaches of AT which indirectly affect the outer optimization by weight perturbations \cite{wu2020adversarial} or weight smoothing \cite{chen2020robust} advocates for further investigation on outer optimization. Based on these observations, we raise a fundamental question regarding outer optimization in AT and attempt to address it in this work:
\begin{adjustwidth}{5pt}{5pt}
{\it Is the conventional MSGD, developed for non-convex optimization in NT, a proper choice for the outer optimization in AT?} {\bf If not}, {\it what modifications are required to make it suitable for the AT setup?}
\end{adjustwidth}

To answer the first question, we empirically evaluate and compare two statistical parameters of gradients, namely expected norm and expected variance, in NT and AT. Both these parameters are known to be major determinants of the performance of MSGD in NT \cite{he2019control,liu2019variance,zhao2020stochastic}. We find that they are notably higher in AT compared to NT. Furthermore, after decaying the learning rate in NT, both the gradient norm and variance deteriorate suggesting convergence to local minima. However, in AT, they escalate after the learning rate decay. These observations highlight substantial disparities between the characteristics of the gradients in AT and NT. Consequently, we argue that MSGD, developed essentially for NT, is not the most proper choice for outer optimization in AT since it is not designed to be robust against high gradient norm and variance. 

Motivated by these observations, the current work attempts to develop an optimization method that is more suitable for AT, \ie less sensitive to the gradient norm and variance. The contributions of the paper are as follows:
\begin{itemize}[leftmargin=*]
    \item We investigate the effect of AT on gradient properties and provide empirical evidence that AT induces higher gradient norm and variance. We argue that this hinders the optimization since the convergence rate of MSGD is highly dependent on the variance of the gradients.
    \item We propose an optimization method tailored specifically for AT, termed ENGM, whose convergence rate is independent of the gradient variance.
    \item 
    We empirically analyze the norm of gradients and provide insightful observations regarding their correlation in DNNs. Harnessing this, we develop a fast approximation to ENGM that significantly alleviates its computational complexity.
    \item Through extensive evaluations and ablation studies, we demonstrate that the proposed optimization technique consistently improves the performance and generalization of the SOTA AT methods. \vspace{-10pt} 
\end{itemize}

\section{Analyzing Outer Optimization in AT}

We first investigate the disparities between the properties of gradients in AT and NT in Section \ref{sec:atvsnt}. Then in Section \ref{sec:revisit}, we draw connections between the observed disparities and poor performance of MSGD in AT by reviewing the previous theoretical analysis on the convergence of MSGD. 
In Section \ref{sec:ENGM}, we describe our proposed optimization technique whose convergence rate is more favorable for AT. Later in Section \ref{sec:accelerating}, we present an interesting observation that enables us to approximate a fast version of the proposed optimization technique.

\subsection{Notations} \label{sec:notations}
Throughout the paper, we denote scalars, vectors, functions,
and sets using lower case, lower case bold face, upper case,
and upper case calligraphic symbols, respectively. We use notation $||\cdot||_p$ for the $\ell_p$-norm and drop the subscript for $p=2$. We employ the commonly used cross-entropy loss as the measure of empirical risk and denote the loss on $i^{th}$ example, $L(F_{\tb}(\x_i), y_i)$, as $L_i$ for the sake of brevity. %

\begin{figure*}[t]
     \centering
         \centering
         \includegraphics[width=\textwidth]{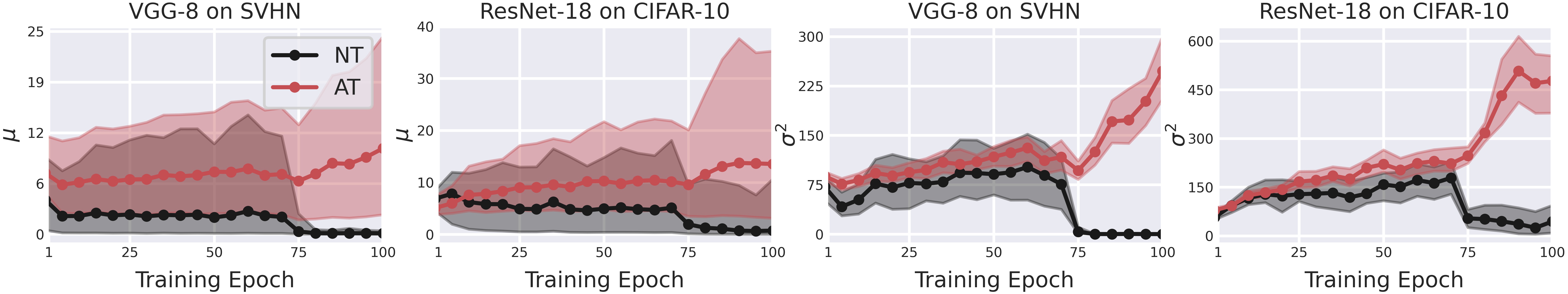}
        \caption{Expected norm ($\mu$) and variance ($\sigma^2$) of gradients during NT and AT. Learning rate is  decayed from $10^{-1}$ to $10^{-2}$ at epoch 75. Note that the norm and variance in AT is higher than NT and escalates after learning rate decay. }
        \label{fig:intro}
\end{figure*}

\subsection{Comparison of Gradient Properties}\label{sec:atvsnt}
We experiment to analyze two statistical parameters of gradients which are major determinants of the performance of MSGD. 
The first parameter is the expected norm of gradients $\mu:=\mathbb{E}_{(\x, y)\sim \mathcal{D}}\big[||\nabla_{\tb} L\big(F_{\tb}(\hat{\x}), y \big)||\big]$, where $\hat{\x}$ is the natural example in NT and the adversarial example in AT. Change in the expected norm directly affects the learning rate, the most important hyperparameter in NT \cite{he2019control,liu2019variance}. The second parameter is the upper bound for the variance of gradients, and is defined as: 
\begin{equation}\label{eq:sigma2}
    \sigma^2 := \sup_{\tb}~~ \mathbb{E}_{(\x, y)\sim \mathcal{D}} \Big[\big|\big|\nabla_{\tb}L\big(F_{\tb}(\hat{\x}), y\big)-\bar{\g} \big|\big|^2 \Big],
\end{equation}
where $\bar{\g}=\mathbb{E}_{(\x, y)\sim \mathcal{D}}\big[ \nabla_{\tb}L\big(F_{\tb}(\hat{\x}), y\big)\big]$. 
It is shown that the convergence of MSGD is $O(\sigma^2)$ \cite{yu2019linear}.    
We roughly estimate both parameters during the training of 
ResNet-18 and VGG-8 on CIFAR-10 and SVHN datasets, respectively. Inner optimization in AT follows the standard setup, \ie $10$ steps of $\ell_\infty$-norm PGD with $\epsilon=8/255$ and step size $\epsilon/4$.

Figure \ref{fig:intro} plots $\mu$ and $\sigma^2$ during $100$ training epochs with the learning rate decay from $10^{-1}$ to $10^{-2}$ at epoch $75$. We observe that the expected norm and variance of gradients is notably higher in AT. 
After learning rate decay, both parameters decrease significantly in NT suggesting the convergence to local minima. However in AT, the expected norm grows and the variance increases drastically. These findings highlight substantial disparities between the characteristics of the gradients in AT and NT. In the next section, we theoretically analyze how these differences can affect the convergence of MSGD. 

\subsection{Revisiting Stochastic Gradient Descent}\label{sec:revisit}
In this part, we analyze the functionally and convergence of MSGD to identify modifications that improves its suitability for the AT setup. The update rule of MSGD at iteration $t$ is as follows:

\begin{subequations}
\begin{equation}\label{eq:msgd_momentum}
    \v_{t+1} = \beta\v_{t} + \dfrac{1}{|\mathcal{I}_t|} \sum_{i\in \mathcal{I}_t} \nabla_{\tb} L_i, 
\end{equation}
\begin{equation}\label{eq:msgd_main}
    \tb_{t+1} = \tb_{t} - \eta \v_{t+1},~~~~~~~~~~~~
\end{equation}
\end{subequations}
where $\eta$ is the learning rate, $\v_{t+1}$ is the Polyak's momentum with the corresponding modulus $\beta$ \cite{polyak1964some}, $\mathcal{I}_t$ is the randomly selected set of indices for the mini-batch with size $|\mathcal{I}_t|$, and $L_i$ is the objective for optimization computed on the $i^{th}$ example. Assuming $F$ has bounded variance of gradients according to Equation \ref{eq:sigma2}, and is smooth in $\tb$, \ie $F_{\tb_1}(\x) \leq F_{\tb_2}(\x) + \langle \nabla_{\tb}F_{\tb_1}(\x), \tb_2-\tb_1\rangle + \tfrac{c}{2} ||\tb_2-\tb_1||^2$, Yu {\it et al.} \cite{yan2018unified,yu2019linear} have shown that the convergence rate of MSGD for non-convex optimization in DNNs is $O(\sigma^2)$. Hence, MSGD is not suitable for tasks with high gradient variance. Intuitively, higher variance implies that the gradients are not aligned with the average gradients which are being used to update the model parameters. This hinders the optimization process since the update is merely favorable for a portion of examples in the mini-batch. 

One alternative to MSGD that is less sensitive to the variance of the gradients is stochastic normalized gradient descent with momentum (SNGM) \cite{zhao2020stochastic}. SNGM is shown to provide better generalization for training with large batch size, \ie another cause of high gradient variance. Concretely, SNGM modifies Equation \ref{eq:msgd_momentum} as:
\begin{equation}\label{eq:ENGM_momentum}
    \v_{t+1} = \beta\v_{t} +  \dfrac{\sum_{i\in \mathcal{I}_t} \nabla_{\tb} L_i}{||\sum_{i\in \mathcal{I}_t} \nabla_{\tb} L_i||} , 
\end{equation}
which limits the gradient norm by normalizing the magnitude of mini-batch gradients and considers only the direction of the average gradient. Zhao {\it et al.} \cite{zhao2020stochastic} have shown that the convergence of SNGM is $O(\sigma)$, and therefore, is more suitable for tasks with induced gradient fluctuations. We also observe in Section \ref{sec:optimcomparison} that SNGM improves the generalization in AT. This suggests that reducing the sensitivity of the optimizer to the gradient variance has a direct impact on the generalization and performance of the task with adversarial gradients. 

\begin{algorithm}[t]
\small
\caption{Fast ENGM}
\label{alg:mlga}
\begin{algorithmic}[1]
\State Initialize $\tau>0$, $\beta_{\gamma}\in[0, 1)$, $\alpha>0$, $\gamma_0=0$, $\gamma_1=1$, Boolean parameter {\it Naive}.
\For{$t=0\dots t_1-1 $}
\State Compute $L_i,~~\forall i \in \mathcal{I}_t$; \Comment{inner optimization}
\State Compute $\mathcal{G}_{\x,t} = \{\nabla_{\x}L_i:i\in \mathcal{I}_t\}$; \Comment{\textcolor{blue}{backprop. $\times 1$}}
\If{mode($t, \tau$)~$=$~$0$ and {\it Naive}~$=$~False}
\State Compute $\mathcal{G}_{\tb,t}=\{\nabla_{\tb}L_i:i\in \mathcal{I}_t\}$; \Comment{\textcolor{blue}{backprop. $\times n$ every $\tau$ iterations}}
\State $\gamma'_1, \gamma'_0 = \text{LinearRegression}(\mathcal{G}_{\x,t}, \mathcal{G}_{\tb,t})$ \Comment{estimate slope and intercept}
\State $\gamma_0\gets \beta_{\gamma} \gamma_0 + (1-\beta_{\gamma})\gamma'_0$, and $\gamma_1\gets \beta_{\gamma} \gamma_1 + (1-\beta_{\gamma})\gamma'_1$;
\EndIf
\State $\hat{w}_i\gets \max(\dfrac{\alpha}{||\gamma_1\nabla_{\x}L_i+\gamma_0||}, 1),~~ \forall i \in \mathcal{I}_t$;
\State Update $\tb$ with MSGD on the reweighted loss $\tfrac{1}{| \mathcal{I}_t|}\sum_{i\in \mathcal{I}_t} \hat{w}_iL_i$ \Comment{backpropagation$\times 1$}

\EndFor

\end{algorithmic}
\end{algorithm}

\subsection{Example-normalized Gradient Descent with Momentum}\label{sec:ENGM}
Although SNGM is less sensitive than MSGD to the variance of gradients, it does not impose any constraint on the variance. Hence, the variance can still become large and impede the optimization. To address this, we introduce a transformation on gradient vectors that bounds the variance of the gradients in the mini-batch and makes the convergence rate of the optimizer independent of the variance. 
\begin{theorem}\label{theorem0}   
For any arbitrary distribution $\mathcal{P}$ of random vectors, applying the transformation $T(\a)=\min(\tfrac{\alpha}{||\a||}, 1)\a$ with $\alpha>0$ bounds the variance of vectors to $4\alpha^2$. \\
(Proof is provided in Section 1 of Supp. material.)
\end{theorem}
We use the transformation in Theorem \ref{theorem0} to bound the variance of the gradients. To this aim, we rewrite Equation \ref{eq:msgd_momentum} as:
\begin{subequations}\label{eq:engm_ur}
\begin{equation}\label{eq:our_updaterule}
    \v_{t+1} = \beta\v_{t} + \dfrac{1}{|\mathcal{I}_t|} \sum_{i\in \mathcal{I}_t}  w_i\nabla_{\tb} L_i, 
\end{equation}
\begin{equation}\label{eq:our_updaterule2}
    w_i = \min\big(\dfrac{\alpha}{||\nabla_{\tb} L_i||}, 1\big), 
\end{equation}
\end{subequations}
where $w_i$ is the normalizing coefficient for $\nabla_{\tb} L_i$, and  $\alpha$ is the maximum allowed norm of gradients. This update rule limits the maximum norm of the gradients on each input example to $\alpha$. Hence, it prevents high magnitude gradients from dominating the updating direction and magnitude in the mini-batch. It might be noted that $\alpha$ scales with the square root of the model size, and larger models require higher values of $\alpha$.  We refer to this approach as {\bf e}xample-{\bf n}ormalized stochastic {\bf g}radient descent with {\bf m}omentum (ENGM). ENGM recovers MSGD when $\alpha \gg 1$. The convergence properties of ENGM is analyzed in Theorem \ref{theorem1}.

\begin{theorem}\label{theorem1} 
Let $A(\tb)$ be the average loss over all examples in the dataset, and assume that it is smooth in $\tb$. 
For any $\alpha>0$ and total iterations of $t_1$, optimizing $A(\tb)$ using ENGM (Equation \ref{eq:engm_ur}) has the convergence of $O(\alpha)$.
(Proof is provided in Section 1 of Supp. material.)
\end{theorem}

Theorem \ref{theorem1} shows that the convergence rate of ENGM is $O(\alpha)$ and is independent of the variance of gradients. Hence, it is suitable for optimizing objectives with high gradient variance. 
Later in Section \ref{sec:optimcomparison}, we empirically validate this and show that the enhanced regularization of ENGM provides better optimization compared to SNGM and MSGD for AT. Despite the intrinsic merits of ENGM, it is computationally expensive since evaluating each $w_i$ requires a dedicated backpropagation and cannot be implemented in parallel. In particular, Equation \ref{eq:engm_ur} requires $|\mathcal{I}_t|$ backpropagation for each mini-batch. 
In the next section, we present an empirical observation on the gradients of DNNs that enables us to estimate $w_i$ and consequently Equation \ref{eq:engm_ur} using merely one additional backpropagation.

\begin{figure*}[t]
     \centering
     \begin{subfigure}[b]{\textwidth}
         \includegraphics[width=\textwidth]{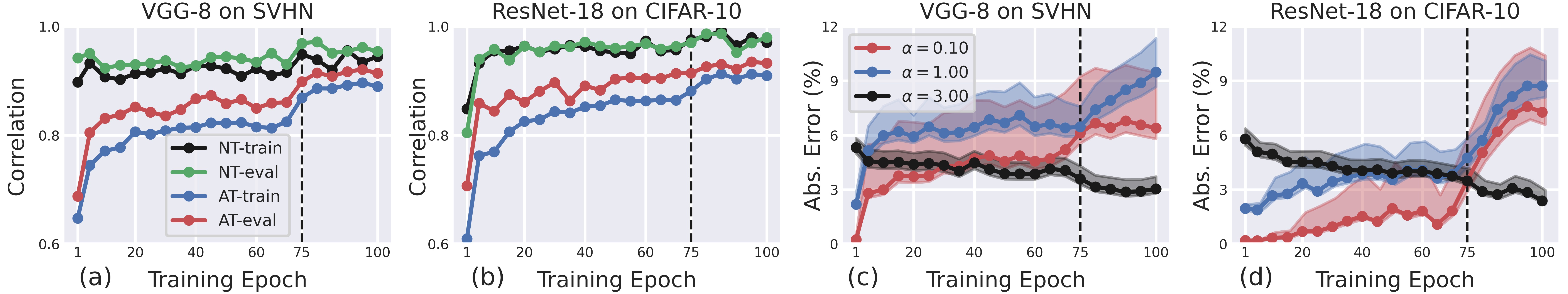}
            \captionlistentry{} 
    \label{fig:corr_a}
     \end{subfigure}
     \begin{subfigure}[b]{0.\textwidth}
         \centering
         \captionlistentry{}
         \label{fig:corr_b}
     \end{subfigure}
     \begin{subfigure}[b]{0.\textwidth}
         \centering
         \captionlistentry{}
         \label{fig:corr_c}
     \end{subfigure}
          \begin{subfigure}[b]{0.\textwidth}
         \centering
         \captionlistentry{}
         \label{fig:corr_d}
     \end{subfigure}
\vspace{-10pt}        \caption{ (a,b): Characterizing the linear correlation between $||\nabla_{x_i}L_i||$ and $||\nabla_{\tb}L_i||$ using Pearson correlation coefficient. 
(c, d): The absolute value of error ($\%$) for estimating $w_i$ using Equation \ref{eq:w_hat}. 
Dashed black line denotes the learning rate decay from $10^{-1}$ to $10^{-2}$. }
\end{figure*}

\subsection{Accelerating ENGM via Gradient Norm Approximation} \label{sec:accelerating}
During our evaluations, we observe an interesting phenomenon that enables us to develop a fast approximation to ENGM.    
Particularly, we observe that the norm of gradients w.r.t. the network parameters, $||\nabla_{\tb}L_i||$, is linearly correlated with the norm of the gradients w.r.t. the input example, $||\nabla_{\x_i}L_i||$. To illustrate this phenomenon, we track both gradient norms on $1,000$ training examples during NT and AT using VGG-8 on SVHN and ResNet-18 on CIFAR-10. We compute Pearson correlation coefficient to measure the correlation between the two norms. Figures \ref{fig:corr_a} and \ref{fig:corr_b} show the correlation coefficient during AT and NT with the model in the evaluation and training modes. We can see that there is a significant correlation between the two norms in DNNs which becomes stronger as the training proceeds. The correlation exists in both the training and evaluation modes of the model, and is slightly affected by the update in the statistics of the batch normalization modules. 

\begin{table*}[t]
\scriptsize
    \centering
    \begin{tabular}{lccccccc}
        \ChangeRT{0.4pt}
    \hline
        Method &  MSGD & MSGD+GNC & SNGM & F-ENGM & N-ENGM & A-ENGM & ENGM \\
        Ex. time (sec./iter) & $0.60$ & $0.61$ & $0.63$ & $5.05$ & $0.75$ & $0.83$ & $5.06$ \\
            \ChangeRT{0.4pt}
    \hline
    \end{tabular}
    \caption{Execution time of the outer optimization methods. Experiments are conducted on an NVIDIA Titan-RTX GPU.}
    \label{tab:ex}
    \vspace{-10pt}
\end{table*}

Harnessing this phenomenon, we can estimate the norm of gradient w.r.t. the network parameters (computationally expensive) using the norm of gradients w.r.t. the inputs (computationally cheap) with a linear approximation as:
\begin{equation}\label{eq:correlation}
    ||\nabla_{\tb}L_i|| \approx \gamma_1 ||\nabla_{\x_i}L_i|| + \gamma_0,
\end{equation}
where $\gamma_0$ and $\gamma_1$ are coefficients for the slope and intercept of the linear estimation, respectively. 
Employing this estimation, we can approximate the functionality of ENGM by a simple modification of the loss on the $i^{th}$ input example, $L_i$, and keeping the popular MSGD as the optimizer. This provides two benefits. First, there is no need to implement a new optimizer enhancing the applicability of the method. Second, the reweighting significantly reduces the computational cost of ENGM. 
To this aim, we use the estimated value for the norm of the gradients w.r.t. the input to normalize the gradients w.r.t. the network parameters indirectly by assigning a weight to the loss function computed on $\x_i$ as $\hat{L}_i:=\hat{w}_i L_i$, where:
\begin{equation}\label{eq:w_hat}
    \hat{w}_i := \max(\dfrac{\alpha}{||\gamma_1\nabla_{\x}L_i+\gamma_0||}, 1).
\end{equation}
Here, optimizing the total loss $\tfrac{1}{|\mathcal{I}_t|}\sum_{i\in \mathcal{I}_t} \hat{L}_i$ using MSGD will approximately recover the functionality of ENGM on $\tfrac{1}{|\mathcal{I}_t|}\sum_{i\in \mathcal{I}_t} L_i$. To analyze the accuracy of estimating $\hat{w}_i$, we measure the average absolute value of the error during the training of the both models in AT and for three different values of $\alpha \in \{0.1, 1.0, 3.0\}$. Figures \ref{fig:corr_c} and \ref{fig:corr_d} visualize the error on two different datasets and network architectures. We observe that the maximum absolute value of error is less than $10\%$ which advocates for the accuracy of estimating $\hat{w}_i$. 
For large values of $\alpha$ the error decreases during the training, while for small values of $\alpha$ the error increases. This points to a trade-off between the estimation error across the training process. It might be noted that the error is computed solely for AT since based on the evaluations in Figures \ref{fig:corr_a} and \ref{fig:corr_d} the correlation is stronger in NT.    

Unlike $\nabla_{\tb}L_i$, $\nabla_{\x}L_i$ can be computed in parallel for a batch of data using a single backpropagation. We consider two approaches for estimating $\gamma_0$ and $\gamma_1$ which result in two variations of ENGM.  In the first approach, referred to as Approximated ENGM (A-ENGM), we evaluate $\nabla_{\tb}L_i$ for a single mini-batch every $\tau$ iterations and use moving average to update the latest estimate. Then for the intermediate iterations, we use the estimate values of $\gamma$ to approximate the norm of gradients using Equation \ref{eq:correlation}. In comparison, A-ENGM reduces the required number of additional backpropagations from $|\mathcal{I}_t|$ (for ENGM) to $1+|\mathcal{I}_t|/\tau$. In practice, we observe that the interval, $\tau$, for estimating $\gamma$ values can be conveniently large as investigated in Section \ref{sec:ablation}. Furthermore, we consider a second approach in which we simply set $\gamma_0=0$ and merge $\gamma_1$ into $\alpha$. We refer to this approach as Naive ENGM (N-ENGM) which solely requires a single additional backpropagation.

\section{Experiments and Analysis}
We evaluate ENGM on three datasets of CIFAR-10, CIFAR-100 \cite{krizhevsky2009learning}, and TinyImageNet \cite{le2015tiny}. Following the benchmark experimental setup for AT \cite{wu2020adversarial,zhang2019theoretically,wang2019improving,cui2021learnable}, we conduct ablation studies and exploratory evaluations on ResNet-18 with $64$ initial channels, originally developed for ImageNet. For SOTA evaluation, we use Wide ResNet-34 with depth factor 10 (WRN-34-10) \cite{zagoruyko2016wide}.

\noindent{\bf Training Setup.} Except for evaluations involving ENGM, all the models are trained using MSGD with momentum $0.9$, weight decay $5\times10^{-4}$ \cite{wu2020adversarial,pang2020bag,gowal2020uncovering}, batch size equal to $128$, and initial learning rate of $0.1$. The learning rate is decayed by $0.1$ at epochs $75$, $90$, and the total number of epochs is set to $120$ unless otherwise noted. The standard data augmentation including random crop with padding size $4$ and horizontal flip is applied for all datasets. All input images are normalized to $[0,1]$. Based on ablation studies in Section \ref{sec:ablation}, we set $\alpha$ for ENGM, A-ENGM, and N-ENGM to $5$, $5$, and $0.5$, respectively. The momentum for A-ENGM is set to $0.7$ based on empirical evaluations. PGD with $10$ steps (PGD$^{10}$),  $\epsilon\!=\!8/255$, and step size $2/255$ is used as the attack to maximize the adversarial loss in $\ell_\infty$-norm ball. As suggested by Rice {\it et al.}  \cite{rice2020overfitting}, during the training we select the model with the highest robust accuracy against PGD$^{20}$ with $\epsilon\!=\!8/255$ and step size $8/(255\times10)$ on a validation set of size $1,000$ as the best model. Only for PGD$^{20}$, we use margin loss instead of cross-entropy due to its better performance in evaluating the robustness of the model \cite{uesato2018adversarial}. 

\noindent{\bf Evaluation Setup.} We evaluate the model against two major attacks. First is the same PGD$^{20}$ used in the training to find the best model. For a more rigorous evaluation of the robust performance, we follow the setup of the recent SOTA defense methods \cite{zhang2019theoretically,wu2020adversarial,zhang2020attacks,cui2021learnable,huang2020self,chen2020efficient,sitawarin2020improving,pang2020bag} and use the benchmark adversarial robustness measure of AutoAttack (AA) \cite{croce2020reliable}. AA  has shown consistent superiority over other white box attacks such as JSMA \cite{papernot2016limitations}, MIM \cite{dong2018boosting}, and CW \cite{carlini2017towards}\footnote{{\it github.com/fra31/auto-attack.}}. Both attacks in evaluations are applied on the test set, separated from the validation set. Maximum norm of perturbation, $\epsilon$, is set to $8/255$ and $128/255$ for $\ell_\infty$-norm and $\ell_2$-norm  threat models. In addition to the robust accuracy, the robust overfitting of the model is computed as the difference between the best and the last robust accuracies (PGD$^{20}$) normalized over the best robust accuracy. All results are the average of three independent runs.

\begin{table}[t]
    \centering\small
    \begin{tabular}{lccccc}
    \ChangeRT{0.4pt}
    \hline
   Optim.  & \multicolumn{4}{c}{Accuracy (\%)} & Overfit.  \\
         Method & Natural &  Best & Last  & AA & (\%) \\\hline
         MSGD & $\boldsymbol{84.70}$ & $50.87$  & $44.15$ & $46.77$ & $13.2$ \\
        MGNC & $83.98$  & $51.88$  & $46.62$ & $47.59$ & $10.1$ \\
        SNGM & $83.73$ & $51.95$ & $46.80$&  $47.75$ & $9.9$  \\
        F-ENGM &$82.91$&$50.05$&$44.04$& $46.54$ & $12.0$ \\ \ChangeRT{0.3pt}\hline
        N-ENGM & $84.36$ & $52.19$ & $48.79$ &$48.06$ & $6.5$ \\  
        A-ENGM & $83.61$ & $52.46$ & $49.75$ & $48.46$ & $5.1$   \\
        ENGM & $83.44$ & $\boldsymbol{53.04}$ &$\boldsymbol{52.76}$& $\boldsymbol{49.24}$ & $\boldsymbol{3.9}$ \\
        \ChangeRT{0.4pt}\hline
    \end{tabular}
    \caption{Comparison of ENGM with MSGD for outer optimization in AT (\textsection \ref{sec:optimcomparison}). `Best' and `Last' refer to the accuracy against PGD$^{20}$ using the best and last checkpoints, respectively. }
    \vspace{-8pt}
    \label{tab:optimcomparison}
\end{table}

\subsection{Comparison of Optimization methods}\label{sec:optimcomparison}
In this section, we evaluate and compare the proposed method with other possible choices for outer optimization in AT. As the fist baseline, we employ the conventional MSGD which is the optimizer in all of the previous AT methods. A popular and well-known trick to bound the gradient norm especially in recurrent neural networks is Gradient Norm Clipping (GNC) \cite{graves2013generating,pascanu2013difficulty}. GNC clips the gradient norm when it is greater than a threshold. This threshold is similar to $\alpha$ in our method. However, instead of bounding the gradient norm on each individual input example, GNC bounds the norm of the average gradients of the mini-batch.  We consider the combination of MSGD with GNC as our second baseline and refer to it as MGNC. The clipping threshold $\alpha$ for MSGD+GNC is set to $25$ based on empirical evaluations. SNGM, discussed in Section \ref{sec:revisit}, is used as the third baseline. For our method, we compare the original ENGM with its accelerated versions, \ie A-ENGM and N-ENGM. The coefficients $\alpha$ and $\tau$ for our methods are set to the best-performing values from Section \ref{sec:ablation}. As an additional baseline, we develop another version of ENGM in which instead of bounding the norm of gradients, we normalize them to the constant value $\alpha$, \ie modifying Equation \ref{eq:our_updaterule} to: $
\v_{t+1} = \beta\v_{t} + \tfrac{1}{|\mathcal{I}_t|} \sum_{i\in \mathcal{I}_t}  \tfrac{\nabla_{\tb} L_i}{||\nabla_{\tb} L_i||}$. We refer to this method as Fixed ENGM (F-ENGM). 

\begin{wrapfigure}[16]{r}{0.5\textwidth}
    \centering
    \vspace{-20pt}
    \includegraphics[width=0.46\textwidth]{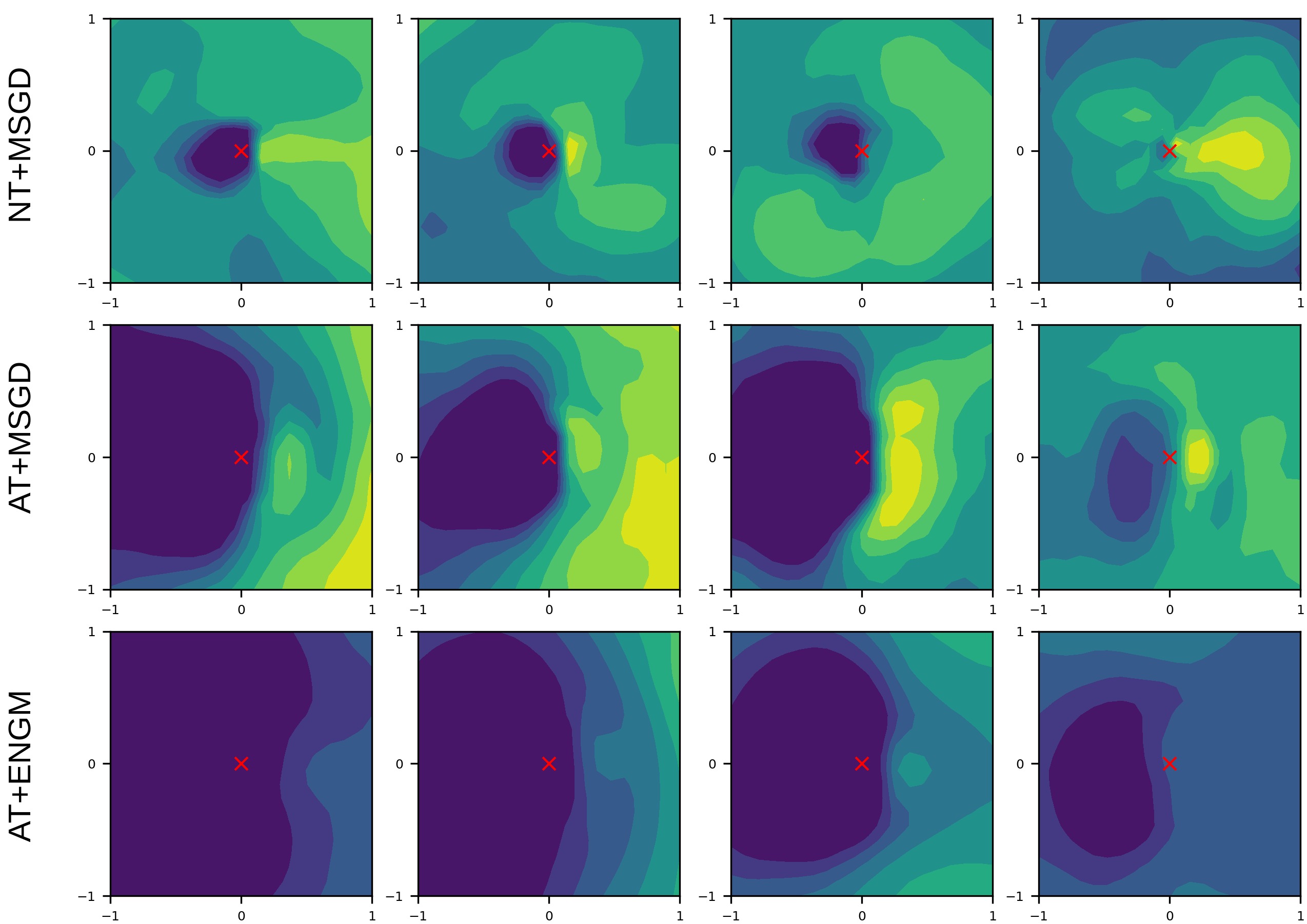}
    \caption{Visualization of the loss landscape on four examples from CIFAR-10 (\textsection \ref{sec:combinationwsota}). The cross mark denotes the input example. Loss level sets are equalized on each column. }
    \label{fig:landscape}
\end{wrapfigure}

\begin{figure*}[t]
     \centering
     \begin{subfigure}[c]{\textwidth}
         \includegraphics[width=\textwidth]{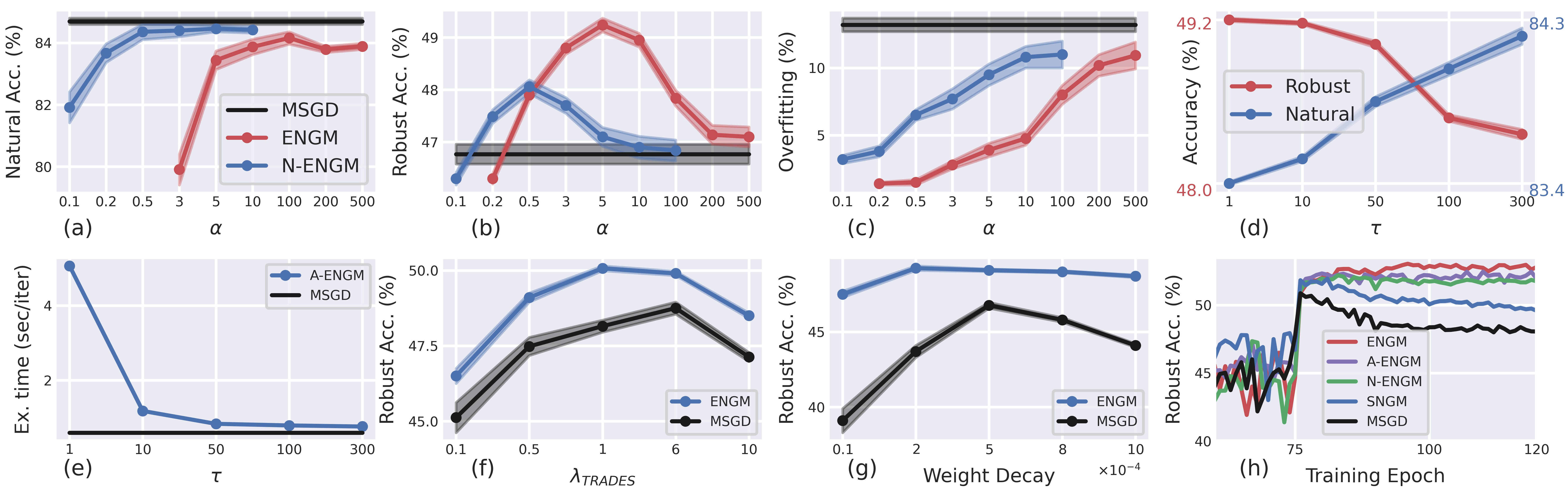}
            \captionlistentry{} 
    \label{fig:ab_a}
     \end{subfigure}
     \begin{subfigure}[b]{0.\textwidth}
         \centering
         \captionlistentry{}
         \label{fig:ab_b}
     \end{subfigure}
     \begin{subfigure}[b]{0.\textwidth}
         \centering
         \captionlistentry{}
         \label{fig:ab_c}
     \end{subfigure}
          \begin{subfigure}[b]{0.\textwidth}
         \centering
         \captionlistentry{}
         \label{fig:ab_d}
     \end{subfigure}
          \begin{subfigure}[b]{0.\textwidth}
         \centering
         \captionlistentry{}
         \label{fig:ab_e}
     \end{subfigure}
     \begin{subfigure}[b]{0.\textwidth}
         \centering
         \captionlistentry{}
         \label{fig:ab_f}
     \end{subfigure}
          \begin{subfigure}[b]{0.\textwidth}
         \centering
         \captionlistentry{}
         \label{fig:ab_g}
     \end{subfigure}

          \begin{subfigure}[b]{0.\textwidth}
         \centering
         \captionlistentry{}
         \label{fig:ab_h}
     \end{subfigure}
\vspace{-10pt}        \caption{(a-g): Ablation studies on $\alpha$, $\tau$, $\lambda_{TRADES}$, and weight decay (\textsection \ref{sec:ablation}). Note that $\alpha$ of ENGM scales to that of N-ENGM with $1/\gamma_1$. Robust accuracy is measured using AutoAttack \cite{croce2020reliable}. (h): learning curves (robust test accuracy) for AT with different outer optimization methods. Results on last 60 epochs are plotted for better visualization of the robust overfitting.  Robust accuracy is measured using PGD$^{20}$. }
\end{figure*}

Table \ref{tab:optimcomparison} presents the results for these comparisons. We can see that the simple GNC enhances robust accuracy providing the same performance as SNGM. These improvements caused by simple modifications further confirms the negative effect of high gradient norm and variance on outer optimization in AT. ENGM consistently improves the robust accuracy over baselines. In addition, robust overfitting in ENGM is significantly lower than other baselines. This suggests that a major cause of robust overfitting in AT is the high fluctuation of gradients and the incompetence of MSGD in addressing it. The learning curves (robust test accuracy) for different optimization methods are depicted in Figure \ref{fig:ab_h}. We observe that after the learning rate decay, the robust performance of ENGM and its variants does not deteriorate which confirms that they alleviate robust overfitting. 
\begin{table}[t]
\setlength{\tabcolsep}{4pt}
    \centering\scriptsize
    \begin{tabular}{lllcccc}    \ChangeRT{0.4pt}\hline        
    
       & AT & Optim. & \multicolumn{3}{c}{Accuracy (\%)}  & Overfit. \\
    &    Method & Method & Natural & PGD$^{20}$ & AA & (\%)\\\hline
         
    \multirow{8}{*}{\rotatebox[origin=c]{90}{CIFAR-10}} &  \multirow{2}{*}{Vanilla} & MSGD & $\boldsymbol{84.70}$ & $50.87$  & $46.77$ & $13.2$  \\ 
         && ENGM & $83.44$ & $53.04$ & $49.24$ & $3.9$ \\
        &\multirow{2}{*}{TRADES} & MSGD & $82.40$ & $50.94$ & $47.85$ & $5.9$  \\
         && ENGM & $82.33$ & $53.46$ & $50.07$ & $3.0$ \\
        &\multirow{2}{*}{MART} & MSGD & $83.68$ & $51.05$ & $48.29$ & $6.1$ \\
         && ENGM & $83.03$ & $53.56$ & $50.48$ & $4.6$  \\
        &\multirow{2}{*}{AWP} & MSGD & $82.98$ & $52.55$ & $50.12$ & $4.6$  \\
         && ENGM & $83.10$ & $\boldsymbol{54.07}$ & $\boldsymbol{51.93}$ & $\boldsymbol{2.7}$ \\    \ChangeRT{0.4pt}\hline         
         
    \multirow{8}{*}{\rotatebox[origin=c]{90}{CIFAR-100}} &  \multirow{2}{*}{Vanilla} & MSGD & $\boldsymbol{57.75}$  & $26.11$  & $24.45$ & $20.9$  \\
         && ENGM & $56.91$  & $28.43$ & $26.60$ & $7.4$ \\
        &\multirow{2}{*}{TRADES} & MSGD & $56.00$ & $29.04$  & $26.93$ & $10.6$\\
         && ENGM & $55.65$ & $30.68$ & $29.20$ & $7.1$ \\
         
        &\multirow{2}{*}{MART} & MSGD & $56.52$ & $29.41$ & $27.18$ & $11.8$ \\
         && ENGM & $56.20$ & $30.89$ & $29.30$ & $8.6$  \\
         
        &\multirow{2}{*}{AWP} & MSGD & $56.22$ & $30.36$ & $28.43$ & $7.3$ \\
         && ENGM & $56.82$ & $\boldsymbol{31.24}$ & $\boldsymbol{30.46}$  &  $\boldsymbol{6.3}$  \\    \ChangeRT{0.4pt}\hline 
         
        \multirow{8}{*}{\rotatebox[origin=c]{90}{Tiny-ImageNet}} &  \multirow{2}{*}{Vanilla} & MSGD & $35.71$ & $7.47$  & $6.92$ & $26.37$ \\
         && ENGM & $29.78$ & $11.29$ & $8.54$ & $10.10$\\
        &\multirow{2}{*}{TRADES} & MSGD & $\boldsymbol{37.26}$ & $14.13$ & $10.95$ & $14.79$ \\
         && ENGM & $36.30$ & $16.88$ & $12.65$ & $8.74$  \\
         
        &\multirow{2}{*}{MART} & MSGD & $37.06$ & $13.79$ & $10.08$ & $15.94$  \\
         && ENGM & $36.53$ & $16.90$ & $12.99$ & $8.20$\\
         
        &\multirow{2}{*}{AWP} & MSGD & $36.13$ & $16.29$ & $13.09$ & $10.67$   \\
         && ENGM & $36.81$ & $\boldsymbol{19.14}$ & $\boldsymbol{16.02}$ & $\boldsymbol{7.97}$ \\    \ChangeRT{0.4pt}\hline 
         
    \end{tabular}
    \caption{Comparison of MSGD and ENGM on different AT methods (\textsection \ref{sec:combinationwsota}). Note that ENGM consistently outperforms MSGD. }
    \label{tab:combwsota}\vspace{-10pt}
\end{table}
The best natural accuracy is provided by MSGD supporting the commonly observed trade-off between the natural and robust accuracies \cite{tsipras2018robustness,zhang2019theoretically}. Table \ref{tab:ex} presents the execution time for the optimization methods. The execution time of ENGM is roughly $8.5\times$ longer than MSGD. However, A-ENGM and N-ENGM achieve notable speed-up and robust performance. As expected, the performance of A-ENGM is between N-ENGM (lower-bound) and ENGM (upper-bound) and is controlled by the estimation interval $\tau$. Hence, we use N-ENGM and ENGM for the major evaluations to clearly compare the two performance bounds.

\subsection{Combination with Benchmark AT Methods} \label{sec:combinationwsota}
In the this section, we incorporate the proposed optimization approaches into the benchmark AT methods including the vanilla method \cite{madry2017towards}, TRADES \cite{zhang2019theoretically}, MART \cite{wang2019improving}, and AWP \cite{wu2020adversarial}. Here, AWP represents the weight perturbation method applied on top of TRADES. The coefficient for the self-distillation loss in TRADES and MART is set to $6$, and the maximum magnitude of weight perturbation for AWP is set to $5\times10^{-3}$. The rest of the training setups are set to the best setup reported by the original papers.  However, the total training epochs for all methods is set to $200$ (learning rate decays by $0.1$ at epochs $100$ and $150$) for the sake of consistency. 

Table \ref{tab:combwsota} presents the results for $\ell_{\infty}$-norm threat model. 
For results on $\ell_2$-norm threat model please refer to Section 2 in Supp. material. 
We observe that ENGM consistently outperforms MSGD on robust performance. The average improvement in robustness against AA is $2.15\%$ and $1.16\%$ in $\ell_\infty$-norm and $\ell_2$-norm, respectively. This suggests that the {\it amount} of perturbation in AT affects the convergence of the outer optimization. Consider the $\ell_2$-norm as the unified metric, the amount of noise in $\ell_\infty$-norm threat model 
is roughly $3\times$ the norm of noise in the counterpart threat model. Combining these results with the evaluations in Figure \ref{fig:intro} advocates that the improvement offered by ENGM over MSGD 
depends on the norm of perturbation. This observation is further investigated in Section \ref{sec:ablation}.

AWP is previously shown to alleviate robust overfitting \cite{wu2020adversarial}. Interestingly, we find that TRADES and MART also reduce the robust overfitting independent of the optimization method. This suggests that the AT method can affect the robust overfitting. ENGM results in the lowest overfitting and consistently surpasses MSGD. On vanilla AT, replacing MSGD with ENGM results in $9.3\%$, $13.5\%$, and $16.2\%$ reduction of overfitting on CIFAR-10, CIFAR-100, and TinyImageNet, respectively. These results advocate that, in addition to the AT method, the outer optimization method also affects the overfitting and limiting the sensitivity of the optimization method to the variance of the gradients can alleviate the robust overfitting. 

As the last evaluation in this part, we visualize the loss landscape on networks optimized by MSGD and ENGM in Figure \ref{fig:landscape}. This figure plots the loss values for the space spanned by the adversarial perturbation (PGD$^{20}$) and random noise, orthogonalized to the perturbation via Gram-Schmidt. We can see that ENGM results in a smoother loss landscape, known as an empirical evidence of the robustness \cite{moosavi2019robustness}. This qualitative analysis further validates the effectiveness of ENGM for outer optimization in AT.  

\subsection{Comparison with SOTA}

Here, we evaluate ENGM in the benchmark of AT, \ie WRN-34-10 on CIFAR-10 dataset \cite{zhang2019theoretically,wu2020adversarial,zhang2020attacks,cui2021learnable,huang2020self,chen2020efficient,sitawarin2020improving}. For training using ENGM, we set $\alpha=10.4$ which is obtained by scaling the best $\alpha$ for ResNet-18 with the factor of $2.08$, square root of the ratio of the total parameters of the two models ($48.2M$ for WRN-34-10 vs. $11.1M$ for ResNet-18). To achieve SOTA performance, we consider AWP as the AT scheme. We train the model for $200$ epochs with learning rate decay by $0.1$ at epochs $100$ and $150$. The rest of the setting is the same as our previous evaluations. Table \ref{tab:wrn3410} presents the results for this experiment. AWP combined with ENGM and N-ENGM surpasses the previous SOTA by $1.28\%$ and $0.94\%$, respectively. This validates the effectiveness of ENGM on large models. We also find that ENGM results in higher natural accuracy on AWP. This suggests that although AWP indirectly improves the outer optimization, its impact is orthogonal to ENGM.

\subsection{Ablation Studies} \label{sec:ablation}
We conduct ablation studies to investigate the impact of hyperparameters on the performance of ENGM and its two variants using ResNet-18 on CIFAR-10. 

\noindent{\bf Impact of $\alpha$:} 
We measure the natural accuracy, robust accuracy (AA), and overfitting versus $\alpha$. We conduct this experiment on ENGM/N-ENGM since they upper/lower bound the performance of A-ENGM.

Figures \ref{fig:ab_a}, \ref{fig:ab_b}, and \ref{fig:ab_c} present the results for these evaluations. As expected, for large values of $\alpha$ all three values converge to that obtained by MSGD. Small values of $\alpha$ can be interpreted as training with a very small learning rate causing both the natural and robust accuracies to drop. Interestingly, we observe that the overfitting decreases significantly for small values of $\alpha$. This confirms that the high variance of gradients in AT negatively affects the functionality of MSGD, \ie ENGM with large $\alpha$. We find that ENGM and N-ENGM achieve their optimal performance on ResNet-18 at $\alpha$ equal to $5$ and $0.5$, respectively. We select these as the optimal values for training the models in other experiments. Note that the optimal value of $\alpha$ is expected to be the same for ENGM and A-ENGM but different for N-ENGM. This is because the formulation of ENGM and A-ENGM is the same except that A-ENGM estimates the norm of gradients every $\tau$ iterations, and setting $\tau=1$ recovers the exact ENGM. However, in N-ENGM, $\alpha$ is scaled by $1/\gamma_1$ according to the discussion in Section \ref{sec:accelerating}. The optimal $\alpha$ is scaled for other networks based on their capacity.

\begin{table}[t]
    \small
    \centering
    \begin{tabular}{llcc}\ChangeRT{0.4pt}\hline 
        Method & Optim. & Nat. Acc. (\%) & AA (\%)  \\\hline
                ATES \cite{sitawarin2020improving} & MSGD & $86.84$ & $50.72$ \\
                BS \cite{chen2020efficient} & MSGD & $85.32$ & $51.12$ \\
                LBGAT \cite{cui2021learnable} & MSGD & $\boldsymbol{88.22}$ & $52.86$ \\
                TRADES \cite{zhang2019theoretically} & MSGD & $84.92$ & $53.08$ \\
                MART \cite{wang2019improving} & MSGD & $84.98$ & $53.17$ \\
                BERM \cite{huang2020self} & MSGD & $83.48$ & $53.34$ \\
        FAT \cite{zhang2020attacks} & MSGD & $84.52$ & $53.51$ \\
        AWP \cite{wu2020adversarial} & MSGD & $85.36$ & $56.17$ \\
        AWP & N-ENGM & $85.40$ & $57.11$ \\
        AWP & ENGM & $86.12$ & $\boldsymbol{57.45}$
        \\\ChangeRT{0.4pt}\hline
    \end{tabular}
    \caption{Comparison of the benchmark robustness on WRN.}
    \label{tab:wrn3410}
\end{table}

\noindent{\bf Impact of $\tau$:}  We conduct experiments to evaluate the role of $\tau$ in AT setup with A-ENGM ($\alpha=5$) as the optimizer and $\tau \in \{1, 10, 50, 100, 300\}$. It might be noted that each epoch in CIFAR-10 consists of $390$ mini-batches of size $128$. Hence, $\tau\!=\!300$ is roughly equivalent to estimating the correlation at the end of each epoch. Figures \ref{fig:ab_d} and \ref{fig:ab_e} present the results for these evaluations. As expected, for small and large values of $\tau$ A-ENGM converges to ENGM and N-ENGM, respectively. For $\tau=50$, obtained robustness is roughly $85\%$ of the robustness obtained by ENGM while the training time is significantly lower ($0.83$ vs. $5.06$) because the extra gradient computation is being performed every $50$ iterations. Furthermore, we can see that $\tau$ controls the trade-off between the natural and robust accuracies.         

\noindent{\bf Perturbation norm:} As an initial exploration in this paper, we observed that AT induces higher gradient norm and variance. We also noticed in Section \ref{sec:combinationwsota} that ENGM seems to outperform MSGD with a larger margin when the magnitude of perturbations is higher. Here, we further analyze the impact of the magnitude of perturbations on the gradient norm and variance induced by AT. This allows us to identify the extent of suitability of MSGD and ENGM for NT and AT. We train models in AT setup with $\ell_\infty$-norm threat model and varied size of perturbation, $\epsilon \in \{0, 2/255, 4/255, 6/255, 8/255, 10/255\}$. Both MSDG and ENGM are utilized for the outer optimization in these evaluations. We measure the average norm and variance of gradients across all training epochs. For a fair comparison, we compute the expected distance to the closest decision boundary as the unified robustness measure: $\rho:=E_{\x}[||\x-\x^*||]$, where $\x^*$ is the closest adversary to $x$ computed using DeepFool \cite{moosavi2016deepfool}. 

Table \ref{tab:pertsize} presents the results for this experiment. In NT (AT with $\epsilon=0$), MSGD provides slightly better performance than ENGM. This is because in NT the norm and variance of gradients are naturally limited. As the $\epsilon$ increases, the expected norm and variance of the gradients also increase. This confirms our initial observation that AT induces higher gradient norm and variance. Consequently as expected, we find that in AT with larger magnitude of perturbations ENGM works better than MSGD.   

\begin{wraptable}[11]{r}{0.50\textwidth}
    \centering\scriptsize
    \vspace{-20pt}
    \begin{tabular}{ccccccc}
    \ChangeRT{0.4pt}
    \hline
     & \multicolumn{6}{c}{Magnitude of Perturbation, $\epsilon$ ($\times \tfrac{1}{255}$)} \\
     & 0 & 2 & 4 &
    6 & 8 & 10 \\\hline
    
    $\mu$ & $4.25$ & $5.10$ &  $6.09$ &$7.73$ &  $10.04$ & $14.21$ \\
    $\sigma^2$ & $118.1$ & $118.7$ & $121.8$ & $141.7$ &  $185.2$ & $253.5$ \\\hline

    $\rho_{\text{MSGD}}$ & $0.33$ & $0.41$ & $0.57$ & $0.93$ & $1.15$ & $1.24$ \\
    $\rho_{\text{ENGM}}$ & $0.30$ & $0.42$ & $0.61$ & $1.08$ & $1.35$ & $1.49$    \\
    \ChangeRT{0.4pt}\hline
    \end{tabular}
    \caption{Analyzing the impact of the perturbation magnitude on gradient properties and final robustness obtained by MSGD and ENGM (\textsection \ref{sec:combinationwsota}). AT with $\epsilon=0$ is equivalent to NT.} 
    \vspace{-30pt}
    \label{tab:pertsize}
\end{wraptable}
\noindent{\bf Sensitivity to hyperparameters:} One intriguing shortcoming of AT is sensitivity to hyperparameter setting. Several works have shown that a slight change in the modulus of the $\ell_2$-norm regularization, \ie weight decay, results in drastic changes in robust performance \cite{pang2020bag,gowal2020uncovering}. Here, we analyze the sensitivity of the proposed optimization method and compare it with that of MSGD. Figure \ref{fig:ab_g} presents the results for this evaluation. We observe that ENGM exhibits significantly less sensitivity to changes in weight decay compared to MSGD. We hypothesis that high weight decay helps MSGD to prevent the bias from input examples with high gradient magnitude. ENGM achieves this goal by explicitly limiting the gradient magnitudes, and thus, is less sensitive to weight decay. We believe this phenomenon calls for more in depth analysis and defer it to future studies.  

\section{Conclusion}
In this paper, we studied the role of outer optimization in AT. We empirically observed that AT induces higher gradient norm and variance which degrades the performance of the conventional optimizer, \ie MSGD. To address this issue, we developed an optimization method robust to the variance of gradients called ENGM. We provided two approximations to ENGM with significantly reduced computational complexity. Our evaluations validated the effectiveness of ENGM and its fast variants in AT setup. We also observed that ENGM alleviates shortcomings of AT including the robust overfitting and sensitivity to hyperparameters.

\clearpage
%
%
\bibliographystyle{splncs04}
\bibliography{egbib}
\end{document}